\documentclass[11pt,a4paper]{article}
\usepackage[hyperref]{acl2019}
\usepackage{times}
\usepackage{latexsym}

\usepackage{url}

\usepackage{amsmath,amssymb}
\usepackage{bm}
\usepackage{mathtools}

\usepackage{xcolor,xcolor-material}
\usepackage{framed}

\usepackage{booktabs}
\usepackage{tabularx}
\usepackage{siunitx}

\usepackage{graphicx}
\graphicspath{ {./images/} }
\usepackage{subcaption}

\usepackage{enumitem}

\aclfinalcopy % Uncomment this line for the final submission
\title{Bridging the Knowledge Gap:\\
Enhancing Question Answering with World and Domain Knowledge
}
\author{Travis R. Goodwin and Dina Demner-Fushman\\
	Lister Hill National Center for Biomedical Communications\\
	U.S. National Library of Medicine\\
	Bethesda, MD, USA
}

\date{}

\begin{document}
\maketitle
\begin{abstract}
%Deep neural networks have demonstrated high performance on many natural language processing (NLP) tasks that can be answered directly from text, and have struggled to solve NLP tasks requiring external (e.g., world) knowledge.
In this paper we present OSCAR (Ontology-based Semantic Composition Augmented Regularization), a method for injecting task-agnostic knowledge from an Ontology or knowledge graph into a neural network during pretraining.
We evaluated the impact of including OSCAR when pretraining BERT with Wikipedia articles by measuring the performance when fine-tuning on two question answering tasks involving world knowledge and causal reasoning and one requiring domain (healthcare) knowledge and obtained $33.3\%$, $18.6\%$, and $4\%$ improved accuracy compared to pretraining BERT without OSCAR and obtaining new state-of-the-art results on two of the tasks.
\end{abstract}

\section{The Problem}
\label{sec:introduction}
``The detective flashed his badge to the police officer.''
The nearly effortless ease at which we, as humans, can understand this simple statement belies the depth of semantic knowledge needed for its understanding: What is a detective? What is a police officer? What is a badge? What does it mean to \textit{flash} a badge? Why would the detective need to flash his badge to the police officer?
Understanding this sentence requires knowing the answer to all these questions and relies on the reader's  knowledge about this world. %: a detective investigates crime, police officers restrict access to the crime scene, and a badge can be a symbol of authority.

As shown in Figure~\ref{fig:copa}, suppose we were interested in determining whether, upon showing the policeman his badge, it is more plausible that the detective would be let into the crime scene or that the police officer would confiscate the detective's badge?
%To answer this question, we would need to leverage our accumulated expectations about the world: 
Although both scenarios are certainly possible, our accumulated expectations about the world suggest it would be very extraordinary for the police officer to confiscate the detective's badge rather than allow him to enter the crime scene.

\begin{figure}[!tbp]
\centering
\small
\begin{framed}
\raggedright
\setlength\fboxsep{2pt}
% \begin{tabularx}{\textwidth}{l@{~}X}
\textbf{Premise:} %&
The \textcolor{MaterialGreen}{detective} flashed his \textcolor{MaterialPink}{badge} to the \textcolor{MaterialBlue}{police officer}.
%\\[.5em]
%\textbf{Question:} %&
What is the most likely \textit{effect}?\\[1em]
\textbf{A:} %&
The \textcolor{MaterialBlue}{police officer} confiscated the \textcolor{MaterialGreen}{detective}'s \textcolor{MaterialPink}{badge}.\\[.5em]
\textbf{B:} %&
The \textcolor{MaterialBlue}{police officer} let the \textcolor{MaterialGreen}{detective} enter the \textcolor{MaterialAmber}{crime scene}.
% \end{tabularx}
\end{framed}
\caption{Example of a question requiring common-sense and causal reasoning \cite{roemmele2011choice}.}
\label{fig:copa}
\end{figure}
Evidence of Grice's Maxim of Quantity \cite{grice1975logic}, this shared knowledge of the world is rarely explicitly stated in text. 
Fortunately, some of this knowledge can be extracted from Ontologies and knowledge bases. 
For example ConceptNet \cite{speer2017conceptnet} indicates that a \textit{detective} is a \textsc{TypeOf} \textit{police officer}, and is \textsc{CapableOf} \textit{finding evidence}; that \textit{evidence} can be \textsc{LocatedAt} a \textit{crime scene}; and that a \textit{badge} is a \textsc{TypeOf} \textit{authority symbol}.

While neural networks have been shown to obtain state-of-the-art performance on many types of question answering and reasoning tasks from raw data \cite{devlin2018bert,rajpurkar2016squad,manning2015computational}, there has been relatively little investigation into how to inject ontological knowledge into deep learning models, with most prior attempts embedding ontological information outside of the network itself \cite{wang2017knowledge}.

In this paper we present a pretraining regularization technique we call OSCAR (Ontological Semantic Composition Augmented Regularization) which is capable of injecting world knowledge and ontological relationships into a deep neural network.
We show that incorporating OSCAR into BERT's pretraining injects sufficient world knowledge to improve fine-tuned performance in three question answering datasets.
The main contributions of this work are:
\begin{enumerate}[nolistsep,noitemsep]
\item OSCAR, a regularization method for injecting ontological information and semantic composition into deep learning models;
\item Empirical evidence showing the impact of OSCAR on two tasks requiring world knowledge, causal reasoning, and discourse understanding even with as few as 500 training example, as well as a task requiring medical domain knowledge; and
%\item Experimental results showing that the same technique used to infer background knowledge about the world can also capture domain-specific knowledge in the case of medical question answering; and
\item An open-source implementation of OSCAR and BERT supporting mixed precision training, non-TPU model distribution, and enhanced numerical stability.
\end{enumerate}

\section{Background}
\label{sec:background}
\textbf{Pretraining.}
The idea of training a model on a related problem before training on the problem of interest has been shown  effective for many natural language processing tasks \cite{dai2015semi,peters2017semi,howard2018universal}.
More recent uses of pretraining adapt transfer learning by first training a network on a language modeling task, and then fine-tuning (retraining) that model for a supervised problem of interest \cite{dai2015semi,howard2018universal,radford2018improving}.
Pretraining in this way has the advantage that the model can build on previous parameters to reduce the amount of information it needs to learn for a specific downstream task. Conceptually, the model can be viewed as applying what it has already learned from the language model task  when learning the downstream task.
%to give it a head start

\textbf{BERT}
(Bidirectional Encoder Representations from Transformers) is a pretrained neural network which has been shown to obtain state-of-the-art results on eleven natural language processing tasks after fine-tuning \cite{devlin2018bert}.
% BERT relies on two pretraining objectives: (1) a variant of language modeling called \textit{Cloze} (originally proposed in \citealt{taylor1953cloze}) where-in 20\% of the words in a sentence are masked and the model must unmask them and (2) a next sentence prediction task where-in the model is given two pairs of sentences and must decide if the second sentence immediately follows the first.
Despite its strong empirical performance, the architecture of BERT is relatively simple: four layers of transformers \cite{vaswani2018attention} are stacked to process each sentence.

\section{The Data}
Incorporating OSCAR into BERT's pretraining requires an embedded ontology and a text corpus.
%
%\textbf{The Ontology.}
% ConceptNet 5 is a semantic network containing relational knowledge contributed to Open Mind Common Sense \cite{singh2002open} and to DBPedia \cite{auer2007dbpedia}, as well as dictionary knowledge from Wiktionary, the Open Multilingual WordNet \cite{singh2002open,fellbaum1998wordnet}, the high-level ontology from OpenCyc\footnote{\url{http://www.cyc.com/opencyc/}}, and knowledge about word associations from ``Games with a Purpose'' \cite{ahn2006games}.
In our experiments we used ConceptNet 5 as our ontology relying on pretrained entity embeddings known as ConceptNet NumberBatch \cite{speer2017conceptnet}.%, which were built using an ensemble of data from ConceptNet, word2vec \cite{mikolov2013efficient}, GloVe \cite{pennington2014glove}, and OpenSubtitles 2016\footnote{\url{http://opus.nlpl.eu/OpenSubtitles-v2016.php}} using retrofitting.
Our text corpus was a 2019 dump of English Wikipedia articles with templates expanded.
%as provided by Wikipedia's Cirrus search engine\footnote{\url{https://www.mediawiki.org/wiki/Help:CirrusSearch}}.
% Preprocessing relied on NLTK's Punkt senter segmenter\footnote{\url{https://www.nltk.org/_modules/nltk/tokenize/punkt.html}} \cite{loper2002nltk}, and the WordPiece subword tokenizer provided with BERT.

%
\begin{figure*}[!htbp]
    \centering
    \includegraphics[width=\textwidth]{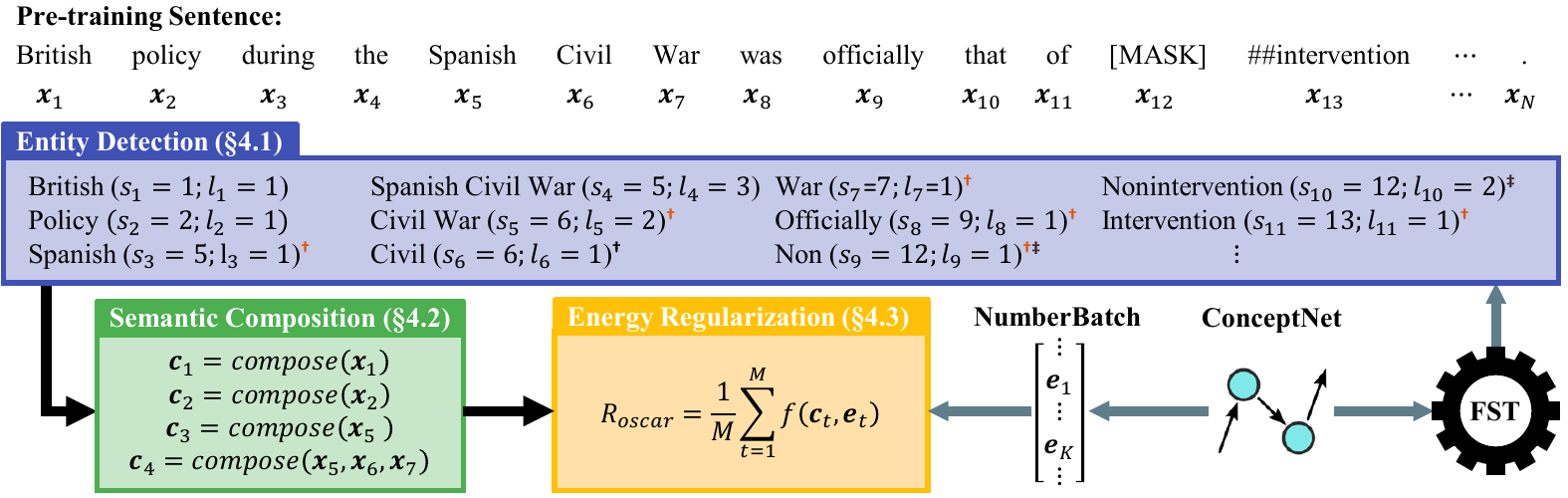}
    \caption{Architecture of OSCAR when injecting ontology knowledge from ConceptNet into BERT where `\textcolor{MaterialOrange900}{${\bm{\dagger}}$}' indicates subsumed entities, `\textcolor{MaterialBrown900}{${\bm{\ddagger}}$}' indicates de-masked entities, $N$ is the length of the input sentence, $M$ is the number of entities detected in the sentence, and $K$ is the number of entities with embeddings in ConceptNet.}
    \label{fig:oscar}
\end{figure*}
\section{The Approach}
\label{sec:methods}
Virtually all neural networks designed for natural language processing represent language as a sequence of words, subwords, or characters.
By contrast, Ontologies and knowledge bases encode semantic information about \textit{entities} which may correspond to individual nouns (e.g., ``badge'') or multiword phrases (``police officer''). 
Consequently, injecting world and domain knowledge from a knowledge base into the network requires \textit{semantically decomposing} the information about an entity into the supporting information about its constituent words. 
% For example, injecting the semantics of ``Spanish Civil War'' into the network requires learning what information the word ``Spanish'' introduces to the nominal ``Civil War'', and what information ``Civil'' adds to the word ``War''.
To do this, OSCAR is implemented using a three step approach illustrated in Figure~\ref{fig:oscar}:

\begin{enumerate}[nolistsep,noitemsep,label={\bfseries Step \arabic*.},leftmargin=*]
    \item entities are recognized in a sentence using a Finite State Transducer (FST);
    \item the sequence of subwords corresponding to each entity are semantically composed to produce an entity-level encoding; and
    \item the average energy between the composed entity encoding and the pretrained entity encoding from the ontology is used as a regularization term in the pretraining loss function.
\end{enumerate}
By training the model to compose sequences of subwords into entities, during back-propagation the semantics of each entity are decomposed and injected into the network based on the neural activations associated with its constituent words.

\begin{figure}[!tbp]
    \centering
    \includegraphics[width=.8\linewidth]{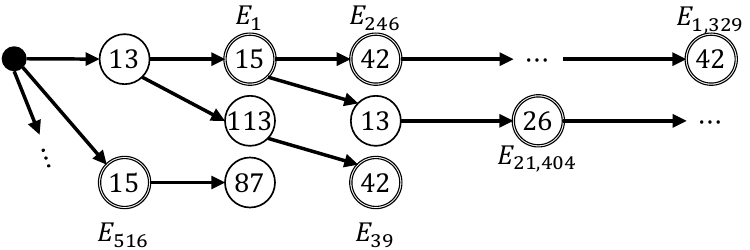}
    \caption{Finite State Transducer (FST) used to detect entities during pretraining; each node corresponds to a word ID, %double circles represent terminal states, 
    and $e_i$ indicates the $i$-th pretrained entity embedding in ConceptNet's NumberBatch.}
    \label{fig:fst}
\end{figure}

\subsection{Entity Detection}
%We designed OSCAR to require as few modification to the underlying host network as possible. 
We recognized entities during training and inference online by (1) tokenizing each entity in our ontology using the same tokenizer used to prepare the BERT pretraining data, and (2) compiling a Finite State Transducer to detect sequences of subword IDs corresponding to entities.
The FST, illustrated in Figure~\ref{fig:fst}, allowed us to detect entities on-the-fly without hard coding a specific ontology or inducing any discernible change in training or inference time.
Formally, let $\bm{X} = \bm{x}_1, \bm{x}_2, \cdots, \bm{x}_N$ represent the sequence of words in a sentence.
The FST processes $\bm{X}$ and returns three sequences: $s_1, s_2, \cdots, s_M$; $l_1, l_2, \cdots, l_M$; and $e_1, e_2, \cdots, e_M$ representing the start offset, length, and the pretrained embedded representation of every mention of any entity in the Ontology.
%
% \paragraph{Entity Subsumption.}
% When detecting entities, it is often the case that multiple entities may correspond to the same span of text.
% As illustrated in Figure~\ref{fig:oscar}, the entity ``Spanish Civil War'' contains the subsumed entities ``Spanish'', ``Civil War``, ``Civil'', and ``War''.
% Likewise, because BERT masks 20\% of the words in each sentence, it is possible for entities to involve masked words.
Note: including or excluding subsumed and de-masked entities (as illustrated in Figure~\ref{fig:oscar}) provided no discernable effect in our experiments.

% \paragraph{Entity De-masking.}
% Because , we evaluated the impact of de-masking words before detecting entities compared with ignoring all entity mentions involving masked words.

\subsection{Semantic Composition}
The role of semantic composition is to learn a composed representation $\bm{c}_1, \bm{c}_2, \cdots, \bm{c}_M$ for each entity detected in $\bm{X}$ such that $\bm{c}_i = \textit{compose}\left(x_{s_i}, x_{s_i + 1}, \cdots, x_{s_i+l_i}\right)$.
As pretraining in BERT is computationally expensive; we considered three computationally-efficient methods for composing words and subwords into entities.

\textbf{Recurrent Additive Networks} (RANs) are a simplified alternative to LSTM- or GRU-based recurrent neural networks that use only additive connections between successive layers and have been shown to obtain similar performance with 38\% fewer learnable parameters \cite{lee2017recurrent}.

Given a sequence of words $\bm{x}_1, \bm{x}_2, \cdots, \bm{x}_L$ we use the following layers to accumulate information about how the semantics of each word in an entity contribute to the overall semantics of the entity:

\vspace{-1em}
\begin{subequations}
\small
\begin{align}
    \bm{\widetilde{m}}_t &= \bm{W}_{m} \bm{x}_t \label{eq:content}\\
    \bm{i}_t &= \sigma\left( \bm{W}_i \left[ \bm{h}_{t - 1}, \bm{x}_t \right] + \bm{b}_i \right) \label{eq:input}\\
    \bm{f}_t &= \sigma\left( \bm{W}_f \left[ \bm{h}_{t -1 }, \bm{x}_t \right] + \bm{b}_f \right) \label{eq:forget}\\
    \bm{m}_t &= \bm{i}_t \circ \bm{\widetilde{m}}_t + \bm{f}_t \circ \bm{c}_{t - 1} \label{eq:memory}\\
    \bm{h}_t &= g\left( \bm{m}_t \right) \label{eq:output}
\end{align}
\end{subequations}
where $[ \bullet ]$ represents vector concatenation,
$\bm{\widetilde{m}}_t$ represents the content layer which encodes any new semantic information provided by word $\bm{x}_t$, 
$\circ$ indicates an element-wise product,
$\bm{i}_t$ represents the input gate, 
$\bm{f}_t$ represents the forget gate, 
$\bm{m}_t$ represents the internal memories about the entity, and 
$\bm{h}_t$ is the output layer encoding accumulated semantics about word $\bm{x}_t$.
We define the composed entity $\bm{c}_i \coloneqq \bm{h}_{s_{i} + l_{i}}$ for the sequence beginning with $\bm{x}_{s_i}$.

\textbf{Linear Recurrent Additive Networks.}
%To further reduce model complexity, we considered 
A second, simpler version of a RAN omits the content and output layers (i.e., Equations~\ref{eq:content} and \ref{eq:output}) and Equation~\ref{eq:memory}. It is updated to depend on $\bm{x}_t$ directly: $\bm{m}_t = \bm{i}_t \circ \bm{x}_t + \bm{f}_t \circ \bm{m}_{t - 1}$.
We define the composed entity $\bm{c}_i \coloneqq \bm{m}_{s_{i}+l_{i}}$ for the sequence of subwords beginning with $\bm{x}_{s_i}$.

\textbf{Linear Interpolation.}
%Finally, we considered the 
The third, simplest form of semantic composition represents the semantics of an entity as an unordered linear combination of the semantics of its constituent words, i.e.: $\bm{c}_i \coloneqq \bm{W}_e  \left(\bm{x}_{s_i} + \bm{x}_{s_i+1} + \cdots +\bm{x}_{s_i + l_i}\right) + l_i \cdot \bm{b}_e$.

\subsection{Energy Regularization}
We project the composed entities into the same vector space as the pretrained entity embeddings from the Ontology, and measure the average energy across all entities detected in the sentence:
\begin{equation}
\small
    \mathcal{R}_{\textsc{OSCAR}} = \frac{1}{M} \sum_{i=1}^M f\left( \bm{W}_p \bm{c}_i + \bm{b}_p, \bm{e}_i \right)
\end{equation}
where $f$ is an \textit{energy function} capturing the energy between the composed entity $c_i$ and the pretrained entity embedding $e_i$.
We considered three energy functions: (1) the Euclidean distance, (2) the absolute distance, and (3) the angular distance.%, which can handle negative values.

\section{Results}
\label{sec:results}
We evaluated the impact of OSCAR on three question answering tasks requiring world or domain knowledge and causal reasoning.

% \begin{table}[!tbp]
%     \centering
%     \begin{tabular}{lrr}
%         \toprule
%          \textbf{Model} & {$\textbf{F}_\bm{1}$} & \textbf{Exact Match} \\
%          \midrule
%          BERT Base      & 86.644054 & 78.722800 \\
%         %  BERT Cirrus    & 88.707203 & 81.419110 \\
%          Oscar          & 86.471826 & 78.552507 \\
%          \bottomrule
%     \end{tabular}
%     \caption{Caption}
%     \label{tab:my_label}
% \end{table}
\textbf{Choice of Plausible Alternatives} 
%a SemEval 2012 shared task,  
(CoPA) presents 500 training and 500 testing sets of two-choice questions and requires to choose the most plausible cause or effect entailed by the premise, as illustrated in Figure~\ref{fig:copa} \cite{roemmele2011choice}.
%The topics of these questions were drawn from two sources:
%(1) personal stories taken from a collection of blogs \cite{gordon2009identifying}; and 
%(2) subject terms from the Library of Congress Thesaurus for Graphic Materials, while the incorrect alternatives were created so as to penalize ``purely associative methods''..

%
\begin{figure}
    \small
    \centering
    \begin{framed}
    \raggedright
    \textbf{Premise:}
    Gina misplaced her phone at her grandparents. 
    It wasn't anywhere in the living room. 
    She realized she was in the car before. 
    She grabbed her dad's keys and ran outside.
    \\[.75em]
    
    \textbf{Ending A:}
    She found her phone in the car.
    \\[.5em]
    
    \textbf{Ending B:}
    She didn't want her phone anymore.
    \end{framed}
    \caption{Example of a Story Cloze question (correct answer is A).}
    \label{fig:cloze}
\end{figure}

\textbf{The Story Cloze Test}
 evaluates story understanding, story generation, and script learning and requires a system to choose the correct ending to a four-sentence story, as illustrated in Figure~\ref{fig:cloze} \cite{mostafazadeh2016corpus}.
In our experiments we used only the 3,744 labeled stories.

\begin{figure}
    \small
    \centering
    \begin{framed}
    \raggedright
    \textbf{Consumer Health Question:}
    Can sepsis be prevented. Can someone get this from a hospital?
    \\[.75em]
    
    \textbf{FAQ A:}
    Who gets sepsis?
    \\[.5em]
    
    \textbf{FAQ B:}
    What is the economic cost of sepsis?
    \end{framed}
    \caption{Example of a Recognizing Question Entailment (RQE) question (correct answer is A).}
    \label{fig:rqe}
\end{figure}

\textbf{Recognizing Question Entailment:}
To overcome the complexity of healthcare questions, Ben Abacha et al. \citeyearpar{abacha2016recognizing} proposed to simplify clinical question answering by answering sub-questions using Recognizing Question Entailment (RQE). 
%, potentially involving family, social, and medical history.
% A proposed solution is to decompose the question into simpler sub-questions which can be more easily answered.
 %Recognizing Question Entailment (RQE, \citealt{abacha2016recognizing}) 
The RQE collection consists of 8,588 training and 302 testing pairs of consumer health questions (CHQs) and frequently asked questions (FAQs) with labels indicating whether answering the FAQ entails answering the CHQ, as illustrated in Figure~\ref{fig:rqe}.

Table~\ref{tab:results} presents the results of BERT when pretrained on Wikipedia with and without OSCAR, the state-of-the-art, and the average performance of different semantic composition methods and energy functions when calculating OSCAR.

\robustify\bfseries
\robustify\itshape

\sisetup{
    round-mode = figures,
    round-precision = 5,
    detect-all = true
}
\begin{table}
\centering
\small
\begin{tabular}{l S[round-precision=3] S S}

\toprule

Model
& \multicolumn{1}{c}{\textbf{CoPA}}
& \multicolumn{1}{c}{\textbf{Cloze}}
& \multicolumn{1}{c}{\textbf{RQE}} 
\\

\midrule
% BERT Base
% && 73.2 %& 52.72459499263623 & 73.96694214876033 & 40.96109839816934 
% && 85.99679315873864
% && 76.49007 & ? & ? & ? \\

BERT
& 55.2 %& 42.27129337539432 & 55.37190082644629 & 34.183673469387754
& 74.2 %& 52.9412 & 75.3138 & 74.8344
& 74.834436 %& ? & ? & ? \\
\\

OSCAR 
& \bfseries 73.6 %& 53.1571 & 64.1803 & 41.4188
& 87.97434526990914 %& ? & ? & ? 
& \bfseries 77.81457 %& \bfseries 77.59197 & \bfseries 68.235296 & 89.92248 \\
\\

SotA
& 71.2
& \bfseries 88.6
& 71.6
\\

\midrule

OSCAR: RAN          
& 60.6 %& 45.9168 & 60.8163 & 36.8812
& 85.88989845002672 %& ? & ? & ?
& \bfseries 77.81457 %& \bfseries 77.59197 & \bfseries 68.235296 & 89.92248 \\
\\

OSCAR: Linear RAN   
& \bfseries 73.6 %& \bfseries 53.1571 & 74.1803 & \bfseries 41.4188 
& \bfseries 87.97434526990914 %& ? & ? & ?
& 75.49669 %& 75.49668 & 65.89595 & 88.372093 \\
\\

OSCAR: Linear
& 72.8 %& 51.6320 & \bfseries 74.359 & 39.5455
& 85.51576696953501 %& ? & ? & ?
& 76.49007 %& 76.87296 & 66.292137 & \bfseries 91.47287 \\
\\

\midrule

OSCAR: Absolute     
& \bfseries 72.0 %& \bfseries 52.2895 & \bfseries 72.541 & \bfseries 40.8776
& 83.43132014965259 %& ? & ? & ?
& 75.49669 %& 76.28204 & 65.02732 & \bfseries 92.248064 \\
\\

OSCAR: Euclidean    
& 60.6 %& 45.9168 & 60.8163 & 36.8812
& 85.88989845002672 %& ? & ? & ?
& 75.49669 %& 75.974023 & 65.363127 & 90.697676 \\
\\

OSCAR: Angular      
& 59.2 %& 46.3902 & 58.9844 & 38.2278 
& \bfseries 86.26402993051844 %& ? & ? & ?
& \bfseries 77.81457 %& \bfseries 77.59197 & \bfseries 68.235296 & 89.92248 \\
\\

\bottomrule

\end{tabular}
\caption{Accuracy when fine-tuning BERT pre-trained on Wikipedia data and pre-trained on Wikipedia data with OSCAR.}
\label{tab:results}
\end{table}

\section{Discussion}
\label{sec:discussion}
% \paragraph{The Impact of World Knowledge.}
%It is clear from Table~\ref{tab:results} that incorporating
OSCAR provided a significant improvement in accuracy for both common sense causal reasoning tasks, indicating that OSCAR was able to inject useful world knowledge into the network.
% We also evaluated the impact of OSCAR on the Stanford Question Answering Dataset (SQuAD), version 1.1 and observed no discern able change in performance (an Accuracy of \num{86.6}\% without and \num{86.5}\% with OSCAR).
% The lack of impact of SQuAD is unsurprising, as the vast majority of SQuAD questions can be answered directly by surface-level information in the text, but it shows that injecting world knowledge with OSCAR does not come at the expense of model performance for tasks that require little outside knowledge.
%
%\textbf{The Impact of Domain Knowledge.}
While less pronounced than the general domain, for the clinical domain, OSCAR provided a modest improvement over standard BERT and both improved over the state-of-the-art.

% \textbf{Entity Subsumption.}
% We evaluated the impact of including subsumed entities when calculating OSCAR, and found it provided, on average, only a minor increase in accuracy ($<1\%$ average relative improvement) at a 10\% increase in total training time. 
% Consequently, we recommend ignoring all subsumed entities.

% \textbf{Entiy De-masking.}
% De-masking entities had little over-all impact on model performance ($<1\%$ average relative improvement), and no discernable effect on training time.
% This may be explained by the fact that Wikipedia sentences are typically much longer than standard English sentences, so the likelihood of an important entity being masked are relatively small.

%\textbf{The role of semantic composition.}
When comparing semantic composition methods, the Linear method had the most consistent performance across both domains; the Recurrent Additive Network (RAN) obtained the lowest performance on the general domain and the highest performance on medical texts, while the Linear RAN exhibited the opposite behavior.
% While this suggests more complex domains require more complex representations of semantic composition, we recommend Linear composition as it exhibits consistent performance and requires 50\% less training time than the RAN and 40\% less than the Linear RAN.

% \textbf{The role of energy functions.}
In terms of energy functions, the Euclidean distance was the most consistent, the Angular distance was the best for the Story Cloze and RQE tasks, and the Absolute difference was the best for CoPA.
The Angular distance (being scale invariant) is least affected by the number of subwords constituting an entity while the Absolute distance is most affected.
Consequently, we believe the Absolute distance was only effective on the CoPA evaluation because the entities in CoPA are typically very short (single words or subwords).
% Consequently, we recommend selecting the energy function based on the average length of entities in the fine-tuning tasks: Angular distance with long entities, Absolute distance with short entities, and Euclidean distance with varied entities.

% Finally, we compared the impact of including and excluding subsumed and masked entities and found that neither resulted in any substantial change in model improvements ($<1\%$ change in accuracy). %, while ignored masked and subsumed entities lead to a 20\% average reduction in training time.

%\subsection{Limitations and Future Work}
% In this study, we only considered ConceptNet as our ontology because we were primarily interested in injecting common-sense world knowledge.
% However, OSCAR is not specific to any Ontology.
% Likewise, we considered only one type of pretrained entity embeddings: ConceptNet NumberBatch \cite{speer2017conceptnet}, despite the availability of other, more sophisticated approaches for knowledge graph embedding including, TransE \cite{bordes2013translating}, TranR \cite{lin2015learning}, TransH \cite{wang2014knowledge}, RESCAL \cite{nickel2011three} and OSRL\cite{xiong2018one}.
In future work we hope to explore the impact of incorporating different Ontologies and knowledge graphs as well as alternative types of entity embeddings \cite{bordes2013translating,lin2015learning,wang2014knowledge,nickel2011three,xiong2018one}.

% \section{Conclusions}
% \label{sec:conclusion}
% In this paper we presented OSCAR (Ontology-based Semantic Composition Augmented Regularization), a learned regularization method for injecting task-agnostic knowledge from an Ontology or knowledge graph into a neural network during pretraining.
% We evaluated impact of including OSCAR when pretraining BERT with Wikipedia articles by measuring the performance when fine-tuning on two question answering tasks involving world knowledge and causal reasoning and one requiring domain (healthcare) knowledge and obtained $33.3\%$, $18.6\%$, and $4\%$ improved accuracy compared to pretraining BERT without Oscar.

\nocite{ole2018parallel}

%\section*{Reproducability}
%All code, data, and experiments are available on GitHub at \textsc{[repository URL withheld]}.

%\section*{Acknowledgments}
% This work was supported by the intramural research program at the U.S. National Library of Medicine, National Institutes of Health.
% This work utilized the computational resources of the NIH HPC Biowulf cluster. (http://hpc.nih.gov)

\bibliographystyle{acl_natbib}
\bibliography{acl2019}

\appendix
\section{Hyper-parameter Tuning}
\subsection{Fine-tuning}
For each fine-tuning task, we used a greedy approach to hyper-parameter tuning by incrementally and independently optimizing:
batch size $\in\allowbreak \left\{\allowbreak 8,\allowbreak 16,\allowbreak 32 \right\}$;
initial learning rate $\in\allowbreak \left\{\allowbreak 1e-5,\allowbreak 2e-5,\allowbreak 3e-5\right\}$;
whether to include subsumed entities $\in\allowbreak \left\{\allowbreak \text{yes},\allowbreak \text{no} \right\}$; and
whether to include masked entities $\in\allowbreak \left\{\allowbreak \text{yes},\allowbreak \text{no} \right\}$.

For CoPA, the Story Cloze task, and RQE we found an optimal batch size of 16 and an optimal learning rate of $2e-5$.
We also found that including subsumed entities and masked was optimal (at a net performance improvement of $<1\%$ accuracy).

\subsection{Pretraining}
We pretrained BERT using a 2019 Wikipedia dump formatted for Wikipedia's Cirrus search engine.\footnote{\url{https://www.mediawiki.org/wiki/Help:CirrusSearch}}
Preprocessing relied on NLTK's Punkt sentence segmenter\footnote{\url{https://www.nltk.org/_modules/nltk/tokenize/punkt.html}} \cite{loper2002nltk}, and the WordPiece subword tokenizer provided with BERT.
We used the vocabulary from BERT base (not large), and a maximums sequence size of 384 subwords, training $64,000$ steps, with an initial learning rate of $2e-5$, and 320 warm-up steps.

\section{BERT Modifications}
We used a modified version of BERT allowing for mixed precision training.
This necessitated a number of minor changes to improve numerical stability around softmax operations.
Training was performed using a single node with 4 Tesla P100s each (multiple variants of OSCAR were trained simultaneously using five such nodes at a time).
Non-TPU multi-gpu support was added to BERT based on Horovod\footnote{\url{https://eng.uber.com/horovod/}} and relying on Open MPI.

\section{State-of-the-Art}
State-of-the-art was determined using the official leader boards maintained by the task organizers.\footnote{CoPA's``leaderboard'' is available at \url{http://people.ict.usc.edu/~gordon/copa.html}.}\footnote{The Story Cloze leaderboard is available at \url{https://competitions.codalab.org/competitions/15333\#results}.}
The RQE evaluation has not concluded; as such, we use the organizers' baseline as the state-of-the-art.

\section{OSCAR Pretraining Recommendations}
When pretraining BERT (or another model) with OSCAR, we make the following recommendations:
\begin{enumerate}[nolistsep,noitemsep]
    \item ignore subsumed entities: including subsumed entities provided only a minor increase in accuracy ($<1\%$ average relative improvement) at a 10\% increase in total training time;
    \item ignore masked entities: De-masking entities had little over-all impact on model performance ($<1\%$ average relative improvement), and no discernible effect on training time;
    \item use linear composition as it exhibits consistent performance and requires 50\% less training time than the RAN and 40\% less than the Linear RAN; and
    \item select the energy function based on the average length of entities in the fine-tuning tasks: Angular distance with long entities, Absolute distance with short entities, and Euclidean distance with varied entities.
\end{enumerate}

\section{Stanford Question Answering Dataset}
We evaluated the impact of OSCAR on the Stanford Question Answering Dataset (SQuAD; \citealt{rajpurkar2016squad}), version 1.1 and observed no discern able change in performance (an Accuracy of $86.6\%$ using BERT pretraining on Wikipedia without OSCAR and $86.5\%$ with OSCAR).
The lack of impact of OSCAR for SQuAD is unsurprising, as the vast majority of SQuAD questions can be answered directly by surface-level information in the text.
However, the lack of impact shows that injecting world knowledge with OSCAR does not come at the expense of model performance for tasks that require little external knowledge.

% \appendix

% \section{Appendices}
% \label{sec:appendix}
% Blah.

% \section{Supplemental Material}
% \label{sec:supplemental}
% bal

\end{document}